\pdfoutput=1

\documentclass[11pt]{article}

\usepackage[]{emnlp2021}

\usepackage{times}
\usepackage{latexsym}
\usepackage{hyperref}

\usepackage[T1]{fontenc}

\usepackage[utf8]{inputenc}

\usepackage{microtype}

\usepackage{mathtools}
\usepackage{amsmath}
\DeclareMathOperator*{\argmax}{arg\,max}

\usepackage{multirow}
\usepackage{hhline}
\usepackage{graphicx}
\usepackage{booktabs}
\usepackage{tikz}
\usepackage{pgfplots}
\usepackage{url}
\usepackage{stfloats}
\usepackage{wrapfig}
\usepackage{numprint}
\usepackage{enumitem}
\pgfplotsset{compat=1.17}

\definecolor{cgreen}{rgb}{0.616, 0.945, 0.616}
\definecolor{lgreen}{rgb}{0.756, 0.960, 0.756}
\definecolor{vlgreen}{rgb}{0.913, 0.988, 0.913}
\definecolor{vvlgreen}{rgb}{0.964, 0.996, 0.964}

\definecolor{cred}{rgb}{0.941, 0.580, 0.580}
\definecolor{lred}{rgb}{0.976, 0.843, 0.843}
\definecolor{vlred}{rgb}{0.984, 0.894, 0.894}
\definecolor{vvlred}{rgb}{0.996, 0.964, 0.964}
\definecolor{darkgreen}{rgb}{0.0, 0.2, 0.13}

\title{A Generative Approach for Mitigating Structural Biases \\ in Natural Language Inference}

\author{Dimion Asael\\
 Technion\\Israel Institute of Technology\\Haifa \\
 \texttt{\href{mailto:dimion@cs.technion.ac.il}{\bf dimion@cs.technion.ac.il}} \\\And
 Zachary Ziegler \\
 SEAS\\Harvard University\\Cambridge, MA \\
 \texttt{\href{mailto:zziegler@g.harvard.edu}{ \bf zziegler@g.harvard.edu}} \\\And
 Yonatan Belinkov \\
 Technion\\Israel Institute of Technology\\Haifa \\
 \texttt{\href{mailto:belinkov@technion.ac.il}{\bf belinkov@technion.ac.il}} \\}

\begin{document}
\maketitle
\begin{abstract}
Many natural language inference (NLI) datasets contain biases that allow models to perform well by only using a biased subset of the input, without considering the remainder features. For instance, models are able to make a classification decision by only using the hypothesis, without learning the true relationship between it and the premise. 
These structural biases lead discriminative models to learn unintended superficial features and to generalize poorly out of the training distribution. 
In this work, we reformulate the NLI task as a generative task, where a model is conditioned on the biased subset of the input and the label and generates the remaining subset of the input. We show that by imposing a uniform prior, we obtain a provably unbiased model. Through synthetic experiments, we find that this approach is highly robust to large amounts of bias. 
We then demonstrate empirically on two types of natural bias that this approach leads to fully unbiased models in practice.
However, we find that generative models are difficult to train and they generally perform worse than discriminative baselines. 
We highlight the difficulty of the generative modeling task in the context of NLI as a cause for this worse performance.
Finally, by fine-tuning the generative model with a discriminative objective, we reduce the performance gap between the generative model and the discriminative baseline, while allowing for a small amount of bias.\footnote{Our code is available at \url{https://github.com/technion-cs-nlp/Generative-NLI}}
\end{abstract}

\section{Introduction}
Natural language processing (NLP) datasets are plagued with artifacts and biases, which allow models to perform the task without learning the desired underlying language capability. 
For instance, in natural language inference (NLI) datasets, 
 models can predict an entailment relationship $y$ from the hypothesis text $H$ alone, without considering the premise $P$ at all \citep{gururangan2018annotation,poliak2018hypothesis}.
 Another identified source of bias includes lexical overlap between $P$ and $H$, which is associated with an entailment prediction \cite{mccoy2020right}. We refer to such biases as \emph{structural biases}, cases where an undesired subset of the input alone incorrectly identifies the label.
Relying on such biases results in poor out-of-distribution (o.o.d) generalization when models are applied to data without bias. Furthermore, models that contain such biases may make surprising predictions when the source of the bias is present, causing problems in critical systems. %

A line of work has attempted to improve the performance on o.o.d datasets by proposing different objective functions \citep[e.g.,][]{utama2020mind,mahabadi2020end}. However, these methods typically still result in a significant gap between the performance in and out of distribution, which indicates that the models are still biased. Table \ref{table:delta} shows this gap, which we term the o.o.d generalization gap ($\Delta$).

In this work, we reformulate classification as a generative task, where the model's task is to generate the remainder features $R$ conditioned on the biased features $B$ and the label $y$. Using Bayes' Rule, we decompose the posterior $p(y\mid B, R)$ into the likelihood $p(R\mid y,B)$ and the prior $p(y\mid B)$. This reformulation lets us control the amount of bias present in the final model. By setting a uniform prior we can obtain a provably unbiased model. %

To assess the extent to which a given model is biased with respect to a specific structural bias, we consider two metrics: the o.o.d generalization gap and the correlation between a model and a biased model $p(y \mid B)$, such as a hypothesis-only or overlap-only model. We first experiment with injecting synthetic bias into a fraction of the training set and evaluating on test sets with and without that bias. We find that the discriminative model's performance decreases as the amount of bias increases, while the generative model maintains similar performance at all bias levels. Moreover, the biased-ness of the discriminative model increases, while the generative model remains unbiased. 

Next, we experiment with two kinds of natural bias: hypothesis-only and overlap bias. We demonstrate that the generative model is unbiased compared to the discriminative baseline as measured by its low o.o.d generalization gap and low correlation with a biased model. 
However, while our approach leads to unbiased models, it performs worse than the discriminative baseline even on o.o.d data. 

We then identify and quantify several causes for the poor performance of the generative model. We show that generative modeling is a more challenging task than discriminative modeling, and that it requires learning a large amount of spurious signal compared to the discriminative model.

Finally, to mitigate the difficulty of the generative modeling task, we fine-tune the generative model with a discriminative objective \citep{lewis2018generative}. While this leaks some bias into the model, the final model matches or surpasses the discriminative baseline while maintaining a relatively small o.o.d generalization gap. %

\begin{table*}[ht]
    \centering
    {
    \begin{tabular}{l c c r c c r}
    \toprule
    \multirow{2}{*}{{}} & \multicolumn{3}{c}{\textbf{SNLI}} &  \multicolumn{3}{c}{\textbf{MNLI}}\\
    \cmidrule(lr){2-4} \cmidrule(lr){5-7}
    {} & \textbf{Test} & \textbf{Hard test} & \multicolumn{1}{c}{\textbf{$\Delta$}} & \textbf{Test} & \textbf{Hard test} & \multicolumn{1}{c}{\textbf{$\Delta$}}\\
    \midrule
    \citet{utama2020mind} & -- & -- & \multicolumn{1}{c}{--} & $82.8\phantom{0}$ & $79.8\phantom{0}$ & $+3.00$\\
    \citet{mahabadi2020end} &  $\boldsymbol{89.57}$ & $\boldsymbol{83.01}$ & $+6.56$ & $\boldsymbol{83.47}$ & $76.83$ & $+6.64$\\
    \citet{sanh2020learning} & -- & -- & \multicolumn{1}{c}{--} & $83.32$ & $\boldsymbol{77.63}$ & $+5.69$\\
    \citet{gururangan2018annotation} (DIIN) & $86.5\phantom{0}$ & $72.7\phantom{0}$ & $+13.8$ & $76.5\phantom{0}$ & $64.4\phantom{0}$ & $+11.1$\\
    \citet{stacey2020avoiding} & $79.39\phantom{0}$ & $69.92$ & $+9.47$ & -- & -- & \multicolumn{1}{c}{--}\\
    Ours (generative, BERT) & $65.53$ & $66.18$ & $\boldsymbol{-0.65}$ & $58.55$ & $57.33$ & $\boldsymbol{+1.22}$\\
    Ours (generative, BART) & $70.58$ & $72.19$ & $-1.61$ & $64.09$ & $65.74$ & $-1.65$\\
    \bottomrule
    \end{tabular}
    }
    \caption{Results on regular and hard (o.o.d) test sets of SNLI and MNLI. Prior work exhibits large o.o.d generalization gaps ($\Delta$), while our generative approach reduces the gap significantly. %
    }
    \label{table:delta}
\end{table*}

To conclude, our contributions are as follows: %
\begin{itemize}[itemsep=2pt,parsep=1pt,topsep=1pt]
    \item We develop a generative modeling approach, which provably eliminates structural biases in natural language understanding tasks.
    \item We demonstrate experimentally on two bias types and different NLI datasets that this approach leads to unbiased models. 
    \item We analyze the strengths and weaknesses of the generative model. 
    \item We show how discriminative fine-tuning improves the generative model, while allowing some bias to leak into the model. %
\end{itemize}

\section{Related Work} \label{sec:related}

\subsection{Biases and Artifacts}
\label{subsec:bias-artifact}
Many natural language understanding (NLU) datasets contain biases or artifacts, superficial features that are associated with a certain label.  Examples include hypothesis-only biases in NLI such as negation words in the hypothesis being correlated with a contradiction label
\citep{poliak2018hypothesis,gururangan2018annotation}. 
Similar one-sided biases have been found in other tasks, including visual question answering (VQA) \citep{agrawal2018don,%
das2019dataset}, reading comprehension \citep{kaushik-lipton-2018-much}, and fact verification \citep{schuster2019towards}. 
Another kind of bias identified in NLI is lexical overlap, which is correlated with an entailment decision in NLI datasets \citep{mccoy2020right}.
We view all these cases as structural biases, cases where the input can be split into two disjoint sets, of the biased features and the remainder features. 

The existence of structural biases in datasets allows models to perform unreasonably well when given access only to the biased features, such as a hypothesis-only model being able to predict entailment without access to the premise. The bias learned by the model manifests in poor o.o.d generalization when evaluated on a test set where the training set correlation between the biased features and a certain label does not hold.

\subsection{Mitigation Strategies}
Common approaches for improving o.o.d generalization combine the main model with a bias model, such as a hypothesis-only model.
For instance, a bias model may be trained adversarially, so that the main model performs worse when the bias model performs well \citep{belinkov2019adversarial,stacey2020avoiding}.
Others use a bias model to modulate the predictions of the main model in different ways \citep{he2019unlearn,mahabadi2020end,utama2020towards}. All these approaches use discriminative models to estimate $p(y \mid P,H)$. Moreover, they typically still result in a gap between the performance in and out of distribution.

In contrast, we propose a novel generative formulation of the NLI task, which leads to an unbiased model, in theory, and in practice.  
\citet{belinkov2019don} also proposed to solve a generative problem, modeling $p(P\mid y,H)$, in order to encourage the model to consider the premise in its predictions.
However, they ended up not using a generative model; rather, they approximated it with discriminative models.
\citet{lewis2018generative} used a generative model for a different task,  VQA, and found it improves generalization from biased training data. While our basic approach is similar, we analyze the generative model more rigorously, investigate the effect of different modeling options, and focus on quantifying the model's bias.

\section{Structural Bias}
\label{section:disc2gen}
Consider the general case of a classification task, for which we are interested in building a model $p_{\theta}(y | X)$ where $y$ is a low-dimensional label and $X$ is an arbitrarily large set of features. The model is trained on an empirical training set $\mathcal{D}=\{(X_i, y_i)\}_{i=1}^N$. The dataset is constructed by humans, and inadvertently contains \textit{structural biases}. We define a structural bias as a case where, if the input $X$ is split into two disjoint sets $X$=$(B, R=X-B)$, the label $y$ can be learned to be reliably predicted given only $B$. For most choices of $B$ this is not a problem, but in some cases, the subset represents an externally imposed constraint that needs to be maintained or an externally imposed understanding of how the model should operate.

This formulation comprises a broad set of biases that are commonly considered. For example, in the NLI task, $X=(P, H)$ where $P$ and $H$ are the premise and hypothesis. If we choose the split $B=H$, we arrive at the hypothesis-only bias. This is an undesirable bias because as humans we know that NLI is impossible if one is only given the hypothesis.

Taking different splits leads to representations of different biases. For instance, we can model the lexical overlap bias under the structural bias framework with the subset $B=P \bigcap H$. NLI models should perform no better than chance when given only the overlapping tokens between $P$ and $H$.

Finally, this formulation extends beyond NLI and NLP to broader biases. For example, if one of the features in $X$ is a protected characteristic $s$ (e.g., gender or race), $B=s$. Then, depending on the task, an undesirable structural bias may exist if a model can learn to predict $y$ given $s$.

We denote these biases as structural biases because they are defined through the structure $B \subset X$, rather than specific known patterns in the data. For example, in the hypothesis-only case, this formulation does not require knowledge about what aspects of the hypothesis allow a hypothesis-only model to predict the label (e.g., negation words), only that somehow the hypothesis alone incorrectly gives a signal about the label. Thus, this type of bias is broader than specific known biases such as the presence of negation words, but narrower than unknown biases because it requires some knowledge of where the bias might be found.

\subsection{Generative Classifiers \textbf{Eliminate} Structural Bias}

Generative classifiers are models that make predictions according to Bayes' Rule. The generative classifier framework provides a principled way of handling structural bias:
\begin{align}
\label{eq:bayes}
    p_{\theta}(y\mid X)&=
    p_{\theta}(y\mid B, R)\\\nonumber
    &=\frac{p_{\theta}(R\mid y, B)p_{\theta}(y\mid B)}{p_{\theta}(R\mid B)} \\\nonumber
    &=\frac{p_{\theta}(R\mid y, B)p_{\theta}(y\mid B)}{\sum_{y'} p_{\theta}(R\mid y', B)p_{\theta}(y'\mid B)}
\end{align}

We emphasize that under this framework, one may %
separately model $p_{\theta}(R\mid y, B)$ and $p_{\theta}(y\mid B)$, but the marginal likelihood must be constructed by marginalizing over the product of those components rather than estimated separately.

Separating the bias component gives explicit control over a given structural bias in the model. Formally, consider the %
ability of any model to predict the label given the bias subset, $p(y \mid B)$, defined by marginalizing out the remainder features:
\begin{align}
\label{eq:real_bias}
    p(y \mid B) = \int p_{\theta}(y \mid R, B) p_{\theta}(R \mid B) \text{d}R.
\end{align}

For a discriminative model this may take any value, but for a generative classifier this becomes:
\begin{align}
\label{eq:uniform}
    p (y\mid B) &= %
    \int p_\theta (y\mid R, B) p_\theta (R\mid B) \text{d}R \\\nonumber
    &=\int p_\theta(R\mid y,B)p_\theta(y\mid B) \text{d}R \\\nonumber
    &=p_\theta(y\mid B) \int p_\theta(R\mid y,B) \text{d}R \\\nonumber
    &=p_\theta(y\mid B).
\end{align}

Therefore, for any given structural bias, the ability of the model to rely on the bias alone, $p(y \mid B)$, can be eliminated in a principled way by training a generative model to learn $p_\theta(R\mid y,B)$ and setting $p_\theta(y\mid B)=\text{Uniform}(\mathcal{Y})$. $R$ and $B$ are collections of tokens, so the actual training process amounts to training a standard autoregressive encoder-decoder model. Predictions are made using Equation~\ref{eq:bayes} at inference time. Unlike other methods, this approach does not require a specific model for $p_\theta(y\mid B)$; it simply requires the \textit{desired} $p_\theta(y \mid B)$, which is often uniform.

\subsection{Measuring Structural Bias}

Typically, debiasing methods are evaluated by measuring the accuracy of the resulting model on a ``hard'' test set, a subset of the test set for which a bias-only model $p(y \mid B)$ predicts the incorrect label. While this captures overall quality, it does not assess the extent to which bias remains. For some applications, the overall quality on non-biased data is a reasonable final objective, but for other applications complete removal of bias is critical.

To quantify the remaining biased-ness of a given model, we consider two metrics: the difference between the accuracy of the model on the standard test set and its accuracy on a ``hard'' set created with respect to the bias in question, which we term the o.o.d generalization gap ($\Delta$), and the correlation ($\rho$) between the predictions of a given model and a fully biased model, i.e., $p(y \mid B)$.

A truly unbiased model will give a similar performance on the original test set and the hard test set, because it cannot rely on the predictive power of $B$ in the original test set even when it is present. Thus low values of $\Delta$ indicate the model is unbiased.

Similarly, %
a model that consistently makes similar decisions to the fully biased model $p(y \mid B)$ in the original test set is likely using only the biased features $B$ as the fully biased model. Therefore, a larger $\rho$ gives additional evidence that a specific structural bias remains in a given model.

\section{Experiments}
\label{subsec:general_experiments}

In all experiments, we estimate $p(R\mid y, B)$ with an encoder-decoder model, with inputs $(y,B)$ and output $R$. 
To condition on $y$, we prefix a label-specific token to $B$.\footnote{\citet{sennrich2016controlling} used a similar approach to control politeness in neural machine translation.} We then train the model as a conditional generative model, by fine-tuning BERT \citep{devlin2019bert} or BART \citep{lewis2020bart} with the standard auto-regressive cross-entropy loss.

At test time, we attach all possible label token to each $B$ to calculate $\argmax_{y' \in \mathcal{Y}} p_\theta (y'|X)$ according to Bayes' Rule in Equation~\ref{eq:bayes}.

\subsection{Synthetic Experiment}
\label{subsec:syn}

To empirically verify the analysis in Section \ref{section:disc2gen}, we construct a synthetic experiment by artificially injecting a hypothesis-only bias into an NLI dataset, similarly to \citet{he2019unlearn}. We use MNLI \citep{williams2018broad},  %
an English NLI dataset, as the base NLI dataset. For each example, we add one of three tokens to the beginning of the hypothesis, each token corresponding to a label. With probability $p$ the token corresponds to the true label and with probability $1-p$ the token is randomly selected from the three labels. The result is that $p$ directly controls the amount of hypothesis-only bias present in the data.

Figure~\ref{graph:delta-rho} (left) shows the results %
when training with synthetic bias in MNLI, for different values of $p$, and evaluating on MNLI dev hard (without synthetic bias), a subset that a hypothesis-only model %
predicts incorrectly. The discriminative model's performance degrades gradually as $p$ increases, while the generative model maintains similar performance. At high levels of $p$, the discriminative model falls below the generative one, indicating that the presence of large amounts of bias precludes the discriminative model from learning the task effectively.
Figure~\ref{graph:delta-rho} (right) shows the two biased-ness metrics, calculated for the generative and discriminative models across a range of $\rho$ values. For each $\rho$, $\Delta$ is calculated from the difference in accuracy for a given model between a version of the dev set with the synthetic bias included as in training, and a version of the dev set with the synthetic bias token randomly chosen for each example. The fully biased model used as the reference when calculating $\rho$ is a model that always selects the label that corresponds with the synthetic bias token prefixed to the hypothesis. According to both metrics, as the bias ratio $p$ increases, the discriminative model quickly becomes significantly biased while the generative model remains entirely unbiased.

\pgfplotsset{every tick label/.append style={font=\small}}

\begin{figure*}
\begin{tikzpicture}
\begin{axis}[
width=\columnwidth-20,
height=\columnwidth-40,
ymin=30,
ymax=80,
xlabel=Bias Ratio,
ylabel=Accuracy,
ytick={30, 40, 50, 60, 70, 80},
enlargelimits = false,
xticklabels from table={Figures/updated_syn.data}{bias-ratio},
xtick=data,
legend style={at={(0.5,0.4)},anchor=north,legend cell align=left,nodes={scale=0.8, transform shape}}]
\addplot[red,dashed,thick,mark=square*] table [y=gen,x=X]{Figures/updated_syn.data};
\addlegendentry{Generative}
\addplot[blue,thick,mark=triangle*] table [y= disc,x=X]{Figures/updated_syn.data};
\addlegendentry{Discriminative}]
\end{axis}
\end{tikzpicture}
\hfill
\begin{tikzpicture}
\begin{axis}[
axis y line*=left,
width=\columnwidth-20,
height=\columnwidth-40,
ymin=-2,
ymax=40,
  xlabel=Bias Ratio,
  ylabel=\textcolor{orange}{$\Delta$},
    ylabel style={rotate=-90},
  ytick={0,10,20,30,40},
  yticklabel style = {color=orange},
  enlargelimits = false,
  xticklabels from table={Figures/delta.dat}{bias-ratio},
  xtick=data,
  legend style={at={(0.33,0.9)},anchor=north,legend cell align=left,nodes={scale=0.8, transform shape,text=black}, legend image post style={black}}
  ]
\addplot[orange,dashed,thick,mark=square*] table [y=delta-gen,x=X]{Figures/delta.dat};
\addlegendentry{Generative}
\addplot[orange,thick,mark=triangle*] table [y=delta-disc,x=X]{Figures/delta.dat};
\addlegendentry{Discriminative}]
\end{axis}
\begin{axis}[
      axis y line*=right,
      axis x line=none,
      width=\columnwidth-20,
      height=\columnwidth-40,
      ylabel=\textcolor{darkgreen}{$\rho$},
      ylabel style={rotate=-90},
      ymin=-0.05,
      ymax=1.0,
      ytick={0.0,0.25,0.5,0.75,1.0},
      yticklabels={0.0,0.25,0.5,0.75,1.0},
      yticklabel style = {color=darkgreen},
      enlargelimits = false,
      legend style={at={(0.4,0.4)},anchor=north,legend cell align=left,nodes={scale=0.8, transform shape}}
    ]
\addplot[darkgreen,dashed,thick,mark=square*] table [y=corr-gen-unbiased,x=X]{Figures/delta.dat};
\addplot[darkgreen,thick,mark=triangle*] table [y=corr-disc-unbiased,x=X]{Figures/delta.dat};]
\end{axis}
\end{tikzpicture}
\caption{Left: Results for models trained with synthetic bias and evaluated on MNLI dev hard without bias. Right: The o.o.d generalization gap ($\Delta$) and the correlation to a bias model ($\rho$) of generative and discriminative models. $\rho$ is calculated on an unbiased test set. %
Refer to Appendix~\ref{ap:corr} for correlations on a biased test set.}
\label{graph:delta-rho}
\end{figure*}
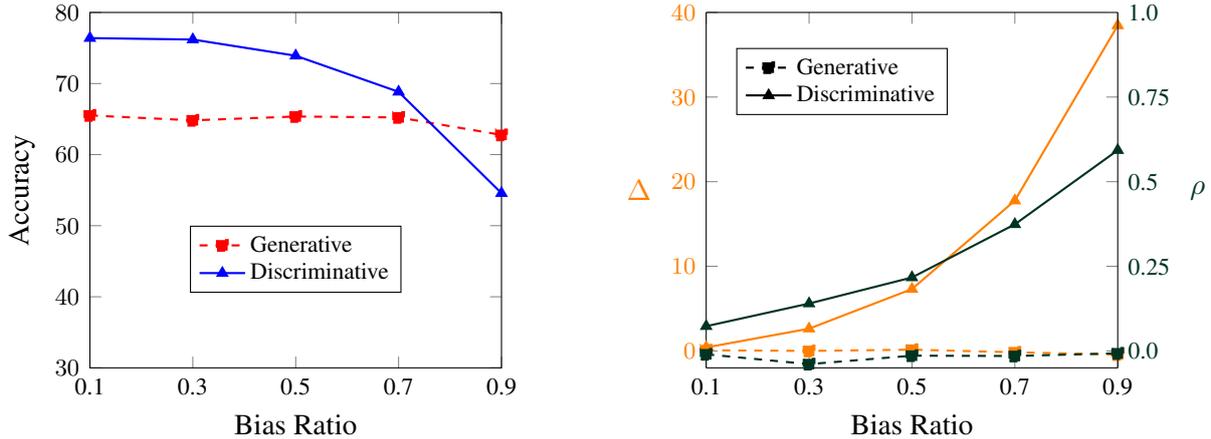

\subsection{Hypothesis-only Bias}
We train our models on the (English) Stanford Natural Language Inference dataset (SNLI; \citealt{bowman2015large}) and on the MNLI dataset, two NLI datasets that are known to contain hypothesis-only biases \cite{poliak2018hypothesis,gururangan2018annotation,tsuchiya2018performance}. We evaluate models on the available in-distribution test sets and on o.o.d test sets that have fewer or no hypothesis-only biases. For SNLI, we use the hard set provided by \citet{gururangan2018annotation}. For MNLI, we use the blind evaluation test\footnote{\url{https://www.kaggle.com/c/multinli-matched-open-evaluation}} and hard test\footnote{\url{https://www.kaggle.com/c/multinli-matched-open-hard-evaluation}} sets for MNLI matched. %

We experiment with two kinds of pre-trained models: a BERT model, which is combined in an encoder-decoder model during the fine-tuning stage, and  BART, a pre-trained encoder-decoder, which is then fine-tuned. For the first kind of model, we used BERT as both an encoder and a decoder. The encoder is a regular BERT model, and the decoder is a BERT model with a language modeling layer, which starts generating from the ``CLS'' token. All models are taken from the Transformers library \citep{wolf2019huggingface}. Both models are fine-tuned with either the baseline discriminative objective or our proposed generative formulation. 

\subsection{Overlap Bias}

Another type of bias that has been demonstrated in the MNLI dataset is lexical overlap bias. \citet{mccoy2020right} demonstrate that, while somewhat uncommon, lexical overlap, subsequence overlap, and constituent overlap between the premise and hypothesis give a strong signal for entailment. Like hypothesis-only bias, this signal comes from peculiarities of the dataset creation process. For a model performing actual NLI, the overlap of words between the hypothesis and the premise should not give any indication of the label. This is emphasized by \citet{mccoy2020right}, as they create a separate label-balanced evaluation set where each example has a high overlap.

To treat overlap bias in the generative formulation, we set $B=P \bigcap H$. Specifically, we concatenate the premise and hypothesis and mask out any tokens that do not appear in both the premise and the hypothesis. The input to the encoder of the generative model is then the label $y$ followed by this partially masked concatenation. For simplicity, the output of the generative model is the unmasked concatenation of $P$ and $H$. In principle, we do not need to output the unmasked tokens, but this simplifies training and remains probabilistically valid.

Because this setup is closely connected to the way the BART model is pretrained, we experiment solely with the BART model for this configuration.

While not traditionally studied in the overlap bias case, we perform the same analysis as in the hypothesis-only bias by constructing a hard %
set for the overlap bias. We train a discriminative model that predicts the label from the masked concatenated premise and hypothesis input, and filter the original MNLI dev set for examples where this discriminative model is incorrect.\footnote{We use the dev set here because the test labels are not available and they are needed for filtering. In this case, we use dev matched for validation and dev mismatched for evaluation.}

\begin{table*}[t]
    \centering
    \resizebox{\textwidth}{!}
    {
    \npdecimalsign{.}
    \nprounddigits{1}
    \begin{tabular}{c l r r r r r r}
    \toprule
     &  & \multicolumn{3}{c}{\textbf{SNLI}} &  \multicolumn{3}{c}{\textbf{MNLI}} \\
    \cmidrule(lr){3-5} \cmidrule(lr){6-8}
    {} & \multicolumn{1}{c}{\textbf{Model}} & \multicolumn{1}{c}{\textbf{Test}} & \multicolumn{1}{c}{\textbf{Hard test}} & \multicolumn{1}{c}{\textbf{$\Delta$}} & \multicolumn{1}{c}{\textbf{Test}} & \multicolumn{1}{c}{\textbf{Hard test}} & \multicolumn{1}{c}{\textbf{$\Delta$}} \\
    \midrule
    \multirow{4}{*}{\rotatebox[origin=c]{90}{BERT}} & Bias-only & $70.82\pm 0.6$ & $32.09\pm 1.7$ & $38.73\pm 1.2$ & $59.77\pm 0.4$ & $34.41\pm 2.3$ & $25.36\pm 2.1$ \\
    {} & Discriminative & $90.49\pm 0.2$ & $80.55\pm 0.3$ & $9.94\pm 0.1$ & $84.08\pm 0.4$ & $76.27\pm 0.3$ & $7.81\pm 0.2$\\
    {} & \begin{tabular}{@{}c@{}}Gen., hyp-only\end{tabular} & $81.42\pm 0.5$ & $61.39\pm 1.4$ & $20.02\pm 0.9$ & $68.5\pm 0.3$ & $52.24\pm 1.4$ & $16.21\pm 1.5$ \\
    {} & Gen., uniform & $65.86\pm 0.3$ & $66.74\pm 0.5$ & $\boldsymbol{-0.88}\pm 0.3$ & $56.98\pm 0.7$ & $54.73\pm 0.2$ & $2.26\pm 0.5$\\
    \midrule
    \multirow{4}{*}{\rotatebox[origin=c]{90}{BART}} & Hypothesis-only & $70.37\pm 0.3$ & $31.61\pm 0.3$ & $38.76\pm 0.1$ & $57.89\pm 2.3$ & $37.83\pm 1.6$ & $20.05\pm 3.9$ \\
    {} & Discriminative & $\boldsymbol{90.78}\pm 0.3$ & $\boldsymbol{81.04}\pm 0.6$ & $9.74\pm 0.3$ & $\boldsymbol{85.67}\pm 0.1$ & $\boldsymbol{78.84}\pm 0.4$ & $6.83\pm 0.4$\\
    {} & \begin{tabular}{@{}c@{}}Gen.,  hyp-only\end{tabular} & $84.36\pm 0.1$ & $67.22\pm 0.8$ & $17.14\pm 0.7$ & $73.85\pm 0.6$ & $60.79\pm 0.6$ & $13.06\pm 0.2$ \\
    {} & Gen., uniform & $70.80\pm 0.2$ & $73.16\pm 0.9$ & $-2.36\pm 0.7$ & $64.22\pm 0.4$ & $64.11\pm 1.0$ & $\boldsymbol{0.11}\pm 0.8$\\
    \bottomrule
    \end{tabular}
    \npnoround
    }
    \caption{Comparison between discriminative baselines and  generative models, with Hyp-only or uniform prior, in the hypothesis-only bias case.}
    \label{table:results}
    \end{table*}

\begin{table}[t]
    \centering
    \resizebox{\columnwidth}{!}
    {
    \npdecimalsign{.}
    \nprounddigits{1}
    \begin{tabular}{l r r r} %
    \toprule
      
    \multicolumn{1}{c}{\textbf{Model}} & \multicolumn{1}{c}{\textbf{Dev}} & \multicolumn{1}{c}{\textbf{Hard dev}} & \multicolumn{1}{c}{\textbf{$\Delta$}} \\ %
    \midrule
    
    Bias-only & $56.32 \pm 0.3$ & $9.37 \pm 8.3$ & $46.95 \pm 8.5$ \\ %
    Disc. & $\boldsymbol{86.44} \pm 0.5$ & $\boldsymbol{79.72} \pm 0.8$ & $6.73 \pm 0.2$ \\ %
    Gen. & $63.67 \pm 1.1$ & $65.56 \pm 0.6$ & $\boldsymbol{-1.88} \pm 0.4$ \\ %
    \bottomrule
    \end{tabular}
    \npnoround
    }
    \caption{Comparison of discriminative and generative models in the lexical overlap bias case. All models are fine-tuned from BART, and the generative model was trained with a uniform prior.}
    \label{table:lexoverlap_results}
    \end{table}

\section{Results}
\subsection{The Generative Model Reduces the o.o.d Generalization Gap}
\paragraph{Hypothesis-only bias}
Table \ref{table:results} shows the results of the proposed generative model and the discriminative baseline in the case of hypothesis-only bias. For the generative model, we show results with either a hypothesis-only prior for $p(y\mid H)$ or a uniform prior. The generative approach with the uniform prior leads to nearly identical accuracy on the i.i.d and o.o.d test sets, that is, unbiased models as measured by low o.o.d generalization gap ($\Delta$ between $-2$ and $3$). %
In contrast, the discriminative model has much larger gaps ($\Delta$ of at least $9$ on SNLI and $7$ on MNLI), meaning that it is a more biased model.  %
The generative model with a hypothesis-only prior also exhibits large generalization gaps, demonstrating the bias leak in this model.  Obviously, a hypothesis-only model is the most biased, with the largest gaps. 

These results also show the advantage of using a pre-trained encoder-decoder (BART) compared to plugging a pre-trained encoder (BERT) and fine-tuning it as an encoder-decoder. While both generative models are unbiased, BART is more amenable to the generative fine-tuning than BERT, with overall better results. For this reason, we only report results with BART henceforth.

\paragraph{Overlap bias}
Table~\ref{table:lexoverlap_results} shows similar results in the case of overlap bias on a hard set w.r.t this bias.\footnote{For results on HANS evaluation sets, see Appendix~\ref{app:hans}.} The generative model exhibits a lower generalization gap ($\Delta$) than the discriminative baseline. As expected, the overlap bias model shows the greatest generalization gap.

\medskip 

While the generative approach leads to unbiased models for both bias types, it also performs significantly worse than the discriminative model, on both in-distribution and o.o.d test sets. We return to this issue in Sections \ref{sec:eval-generations} and \ref{sec:finetune}.

\begin{table}[t]
    \centering
    \begin{tabular}{l r r r}
    \toprule
        \multicolumn{1}{l}{Model} & \multicolumn{1}{c}{Hyp-SNLI} & \multicolumn{1}{c}{Hyp-MNLI} & \multicolumn{1}{c}{Overlap} \\
    \midrule
         Disc. & $0.271$ & $0.223$ & $0.171$ \\
         Gen. & $-0.025$ & $-0.009$ & $-0.043$ \\ 
         Majority & $0.005$ & $0.055$ & $0.016$ \\ 
         Uniform & $-0.018$ & $-0.006$ & $0.007$ \\
    \bottomrule
    \end{tabular}
    \caption{Correlations of discriminative, generative, majority, and uniform models with bias models, on hyp-only (on SNLI/MNLI) and overlap bias (on MNLI).}
    \label{table:correlations}
\end{table}

\subsection{The Generative Model is Uncorrelated with a Bias Model}
\label{subsec:corr}
Table~\ref{table:correlations} shows correlations $\rho$ of the generative model and the discriminative baseline with a bias-only model. 
In the hypothesis-only case, the models were trained on SNLI or MNLI and %
correlations were measured on predictions on SNLI test or MNLI dev mismatched, respectively. In the overlap case, the models were trained on MNLI and %
correlations were measured on MNLI dev mismatched.

In both bias types, the discriminative model predictions are much more correlated with the bias models than the predictions of the generative models. In fact, the correlations of the generative models are as low as those of a majority model or a uniform model, which is unbiased by construction.

\begin{table*}[t]
  \centering
    \resizebox{\textwidth}{!}
  {
  \begin{tabular}{c|c|c}
  \toprule
  \textbf{Label} & \textbf{Hypothesis} & \textbf{Generated premise}\\
  \midrule
     contradiction & {} & an elderly woman is sitting on a bench with her legs crossed and [...] \\ %
     entailment & the woman is young & a young woman in a black shirt and jeans is walking down the street\\
     \textbf{neutral} & {} & a woman in a red shirt is sitting on a bench with a bag in her lap\\
  \bottomrule
  \end{tabular}
  }
  \caption{Examples of premises generated by the model from <label-hypothesis> pairs. Each hypothesis was originally in the SNLI dataset with the premise \textit{"A woman with a green headscarf, blue shirt and a very big grin"}. Ground truth labels for the hypotheses w.r.t the original premise are shown in \textbf{bold}.}
  \label{table:gen-premise}
 \end{table*}
 
 \begin{table*}[t]
    \centering
    \resizebox{\columnwidth * 2}{!}
    {
    \begin{tabular}{l|l|c}
    \toprule
    \multicolumn{1}{c}{\textbf{Premise}} &  \multicolumn{1}{c}{\textbf{Hypothesis}} & \textbf{Label}\\
    \midrule
         \colorbox{vlgreen}{\strut a}woman in \colorbox{vvlgreen}{\strut a black shirt}\colorbox{cred}{\strut looking}\colorbox{vlred}{\strut at}\colorbox{lred}{\strut a}\colorbox{vlred}{\strut bicycle}. & \begin{tabular}{@{}l@{}}\colorbox{vvlgreen}{\strut a}\colorbox{vlred}{\strut woman}\colorbox{lgreen}{\strut dressed}\colorbox{vlgreen}{\strut in}\colorbox{lgreen}{\strut black}\\\colorbox{vvlred}{\strut shops}\colorbox{lgreen}{\strut for}\colorbox{vvlgreen}{\strut a}\colorbox{lgreen}{\strut bicycle}.\end{tabular} & entailment\\
         \midrule
         \begin{tabular}{@{}l@{}}a\colorbox{lgreen}{\strut black}\colorbox{cgreen}{\strut man}in a white uniform makes a spectacular \\ reverse slam dunk to the crowd ' s amazement.\end{tabular} & \colorbox{vlred}{\strut the}man\colorbox{vlred}{\strut is}\colorbox{cgreen}{\strut asian} & contradiction\\
    \bottomrule
    \end{tabular}
    }
    \caption{Gradient attributions example. Green/red show positive/negative attributions.}
    \label{table:attributions}
  \end{table*}

\section{Evaluating Generated Premises}
\label{sec:eval-generations}
So far, we have only used the generative model to score existing examples (with teacher forcing), conditioned on the label and the biased features. In this section, we evaluate the quality of its generations when decoding without constraints. For the experiments here, we consider the hypothesis-only bias and evaluate the quality of the generative model in generating premises. We use a BART model trained on SNLI and generate premises for all hypotheses in the test set. 

To evaluate how well our model can generate premises, we used two metrics: BLEU \citep{papineni2002bleu} of the generated premises w.r.t gold premises, to measure the generation quality (higher is better), and self-BLEU \citep{zhu2018texygen} to measure the diversity of the generations (lower is more diverse).
We report a BLEU value of $0.1078$, indicating that the model is not very good at generating premises. We report self-BLEU of $0.8032$ for the generated premises compared to $0.5875$ for the original premises, suggesting that the generated premises are less diverse. Table~\ref{table:gen-premise} also shows examples where, given different hypotheses, the model generates very similar premises.

A possible explanation for the difficulty of the generative task may be found in the nature of NLI examples in common datasets.
In many cases, the relationship is determined by a small number of words in the premise and hypothesis pair.  To quantify this, we measured the number of words highlighted as explanations in the e-SNLI dataset \citep{camburu2018snli} and found that less than 21\% of words in the premise are highlighted on average.\footnote{These are conservative results, after filtering out premises that did not include any highlighted words.} 
This pattern is reflected also in decisions made by NLI models. By applying gradient attributions,\footnote{We computed attributions with Integrated Gradients \citep{sundararajan2017axiomatic} using Captum \citep{kokhlikyan2020captum}.} we found that more than 70\% of the premise words have low attributions values (between $-0.1$ to $0.1$), with fewer than 6\% of the words having absolute values greater than $0.3$. This shows that only a small number of words had any significant effect on the model predictions. 
Table~\ref{table:attributions} shows a qualitative example of this behavior. Finally, this pattern is also reflected in the generations produced by the generative model, as demonstrated in Table \ref{table:gen-premise}.

\section{Discriminative Fine-tuning} \label{sec:finetune}
The analysis in Section \ref{sec:eval-generations} suggests that the central limitation of the generative model is that at training time there is no indication of the model's downstream use as a classifier. The model is rewarded at training time for devoting significant capacity to modeling the full high-dimensional distribution of $R$, even when large parts of that distribution are unimportant for making downstream predictions.

To help the generative model in such cases, we experiment with an additional fine-tuning step in which we directly optimize for predictive performance. Specifically, for the fine-tuning step we construct the discriminative distribution using Bayes' Rule in Eq. \ref{eq:bayes} and use it at \textit{training time} by minimizing the label cross-entropy loss:
\begin{align}
\label{eq:disc-obj}
    \mathcal{L}_{ft}
    &=-\sum_{i=1}^N \log p_{\theta}(y_i\mid B_i, R_i) \\
    &=-\sum_{i=1}^N \log \frac{p_{\theta}(R_i\mid y_i, B_i)p_{\theta}(y_i\mid B_i)}{\sum_{y'} p_{\theta}(R_i\mid y', B_i)p_{\theta}(y'\mid B_i)}.\nonumber 
\end{align}

Using this objective requires a choice of $p_\theta(y \mid B)$. We explore the impact of different choices for this distribution in the Appendix, but we find that using a pretrained and frozen $p_\theta(y \mid B)$ during the fine-tuning step works best. We hypothesize that this setup allows the generative component $p_{\theta}(R\mid y, B)$ to ignore as much bias as possible.

At inference time, as we are interested in ignoring the bias, we take the fine-tuned generative component $p_\theta(R \mid y, B)$ and perform inference in the same way as before, using Bayes' Rule with a uniform prior.

\begin{table*}[t]
    \centering
    \resizebox{\textwidth}{!}
    {
    \begin{tabular}{l r r r c r r r c}
    \toprule
     & \multicolumn{4}{c}{\textbf{SNLI}} &  \multicolumn{4}{c}{\textbf{MNLI}}\\
    \cmidrule(lr){2-5} \cmidrule(lr){6-9}
    \multicolumn{1}{c}{\textbf{Model}} & \multicolumn{1}{c}{\textbf{Test}} & \multicolumn{1}{c}{\textbf{Hard test}} & \multicolumn{1}{c}{\textbf{$\Delta$}} & \multicolumn{1}{c}{\textbf{$\rho$}} & \multicolumn{1}{c}{\textbf{Test}} & \multicolumn{1}{c}{\textbf{Hard test}} & \multicolumn{1}{c}{\textbf{$\Delta$}} & \multicolumn{1}{c}{\textbf{$\rho$}}\\
    \midrule
    Disc. & $\boldsymbol{90.78}\pm 0.3$ & $81.04\pm 0.6$ & $9.74\pm 0.3$ & $0.27$ & $\boldsymbol{85.67}\pm 0.1$ & $\boldsymbol{78.84}\pm 0.4$ & $6.83\pm 0.4$ & $0.22$\\
    Gen. & $86.30\pm 0.4$ & $\boldsymbol{82.20}\pm 0.3$ & $\boldsymbol{4.09}\pm 0.1$ & $\boldsymbol{0.09}$ & $79.66\pm 1.5$ & $76.45\pm 0.7$ & $\boldsymbol{3.21}\pm 1.3$ & $\boldsymbol{0.07}$\\
    \bottomrule
    \end{tabular}
    }
    \caption{Fine-tuned model results for hypothesis-only bias. Disc.\ is the discriminative baseline, Gen.\ is a generative model fine-tuned with a discriminative objective.}
    \label{table:fine-tune-results-hyp}
\end{table*}

The adjusted training procedure is composed of the following steps: 
1) Train a discriminative prior model, $p_\theta(y\mid B)$, freeze the weights. 2) Train a generative model, $p_\theta(R \mid y, B)$, as in Section \ref{subsec:general_experiments}. 3) Fine-tune the model using Eq. \ref{eq:disc-obj}, using the pretrained $p_\theta(y\mid B)$. 4) Test the model using Eq.~\ref{eq:bayes} with a uniform prior.

\subsection{Results}
Tables~\ref{table:fine-tune-results-hyp} and~\ref{table:fine-tune-results-overlap} show the results of the fine-tuning pipeline. 
The fine-tuned generative models achieve smaller o.o.d generalization gaps ($\Delta$) and correlations to the biased models ($\rho$) than the discriminative baselines. The fine-tuned models are also significantly better than purely generative models in terms of o.o.d performance, at the expense of slight bias leakage (higher $\rho$ compared to pure generative models in Table \ref{table:correlations}). 
In the case of hypothesis-only bias, the fine-tuned models match or surpass the results of the discriminative baselines on the o.o.d sets. In the overlap bias case, the fine-tuned model does not match the discriminative model on the o.o.d set, but it narrows the gap. 

The above results were obtained using a bias model prior in the fine-tuning step and a uniform prior at inference time. This was the strategy that achieved the lowest generalization gap ($\Delta$) on the dev set while outperforming the discriminative baseline. See  Appendix~\ref{app:ablations} for an ablation study of additional options.

\begin{table}[t]
    \centering
    {
    \begin{tabular}{p{1.1cm} c c c r}
    \toprule
    \multicolumn{1}{l}{\textbf{Model}} & \multicolumn{1}{c}{\textbf{Dev}} & \multicolumn{1}{c}{\textbf{Hard dev}} & \textbf{$\Delta$} & \multicolumn{1}{c}{\textbf{$\rho$}}\\
    \midrule
    Disc. & $\boldsymbol{86.44}$ & $\boldsymbol{79.72}$ & $6.73$ & $0.171$\\
    Gen. & $79.87$ & $74.98$ & $\boldsymbol{4.89}$ & $\boldsymbol{0.106}$ \\
    \bottomrule
    \end{tabular}
    }
    \caption{Fine-tuned model results for overlap bias on MNLI mismatched dev set.}
    \label{table:fine-tune-results-overlap}
\end{table}

\section{Conclusion}

Structural biases are common in various NLI datasets and are a major obstacle when trying to create robust systems for this task.
We proposed a generative approach for NLI, which leads to unbiased models. We demonstrated that our generative models are robust to large amounts of bias and perform equally well in and out of distribution. 
This comes, however, with a trade-off, where the generative models perform worse than discriminative baselines. We investigated reasons for the difficulty of training generative NLI models, highlighting the large output space of generating sentences, as opposed to identifying a small subset of words that are often sufficient for solving the task.  We showed how to mitigate this problem by fine-tuning the generative model with a discriminative objective. 
Our work lays down a novel formulation for the NLI task, which may be applied to many other natural language understanding tasks. Future work can examine other kinds of bias and different tasks, as well as new ways to improve the generative model.

\section*{Acknowledgements}
We thank Alexander Rush for helpful discussions. This research was supported by the ISRAEL SCIENCE FOUNDATION (grant No. 448/20) and by an Azrieli Foundation Early Career Faculty Fellowship. Y.B.\ was supported by the Viterbi Fellowship in the Center for Computer Engineering at the Technion.

\bibliography{anthology,custom}
\bibliographystyle{acl_natbib}

 \clearpage 
\appendix

\section{Appendix}
\label{sec:appendix}

\subsection{HANS evaluation sets} \label{app:hans}
Given that we study overlap bias, it is natural to evaluate the generative classifier approach on the HANS evaluation set \citep{mccoy2020right}. The three HANS evaluation sets are constructed to be label balanced, while containing exclusively examples of significant amount of overlap. Three specific types of overlap are considered: lexical overlap, subsequence overlap, and constituent overlap.
See \citet{mccoy2020right} for more details.

Table \ref{table:hans_results} shows the accuracies of the generative classifier and previous results from the literature, reported by \citet{utama2020mind}. 
In general, the accuracies for the generative classifier are low. We hypothesize that this is due to the fact that the examples in the HANS evaluation set are significantly out of distribution compared to the training set, with respect to the amount of overlap between premise and hypothesis. In the training set, sentences often have 20 or 30 tokens with only 1 or 2 token overlaps. In the HANS set, sentences are shorter and all but 1 or 2 tokens overlap. This makes the input significantly more out of domain for the generative classifier only, which is used to seeing many mask tokens in the input and in the HANS set sees almost no mask tokens.

\begin{table}[ht]
    \centering
    {
    \npdecimalsign{.}
    \nprounddigits{1}
    \begin{tabular}{p{2.85cm} r r r}
    \toprule
      
    \multicolumn{1}{c}{\textbf{Model}} & \multicolumn{1}{c}{\textbf{Lex.}} & \multicolumn{1}{c}{\textbf{Subseq.}} & \multicolumn{1}{c}{\textbf{Const.}} \\
    \midrule
        
    Hypothesis-only & 48.2 & 48.7 & 50.4 \\
    Discriminative & 80.7 & 55.5 & 66.3 \\
    \midrule
    Learned-mixin  & 77.5 & 54.1 & 63.2 \\
    PoE  & 72.9 & 65.3 & 69.6 \\
    Conf. reg.  & 73.3 & 66.5 & 67.2 \\
    \midrule
    Generative & 50.7 & 57.7 & 53.2 \\
    \bottomrule
    \end{tabular}
    \npnoround
    }
    \caption{Discriminative and generative models evaluated on the three HANS evaluation sets.}
    \label{table:hans_results}
    \end{table}

\subsection{Correlations} \label{ap:corr}
Figure~\ref{graph:rho-biased} shows the correlations of generative and discriminative models to a bias model under different bias ratios in the synthetic bias case. Here the correlations are calculated on a biased test set, while in Section~\ref{subsec:corr} they were calculated on an unbiased test set. The pattern is the same: the discriminative model is become more biased (higher $\rho$) as the bias ratio increases, while the generative model remains unbiased (small $\rho$). 

\pgfplotsset{every tick label/.append style={font=\small}}
\begin{figure}[ht!]
\resizebox{\columnwidth}{!}{
\begin{tikzpicture}
\begin{axis}[
      xlabel=Bias Ratio,
      width=\columnwidth-3.2,
      height=\columnwidth-15,
      ylabel=\textcolor{black}{$\rho$},
      ylabel style={rotate=-90},
      ymin=-0.05,
      ymax=1.0,
      ytick={0.0,0.25,0.5,0.75,1.0},
      yticklabels={0.0,0.25,0.5,0.75,1.0},
      yticklabel style = {color=black},
      enlargelimits = false,
      xticklabels from table={Figures/delta.dat}{bias-ratio},
      xtick=data,
      legend style={at={(0.5,0.7)},anchor=north,legend cell align=left,nodes={scale=0.8, transform shape}}
    ]
\addplot[green,thick,dashed,mark=square*] table [y=corr-gen,x=X]{Figures/delta.dat};
\addlegendentry{Generative}
\addplot[darkgreen,thick,mark=triangle*] table [y=corr-disc,x=X]{Figures/delta.dat};
\addlegendentry{Discriminative}
]
\end{axis}
\end{tikzpicture}
}
\caption{The correlation to a bias model ($\rho$) of generative and discriminative models under different bias ratios. $\rho$ is calculated on a biased test set, so that each model used the same bias ratio at training and inference time.}
\label{graph:rho-biased}
\end{figure}
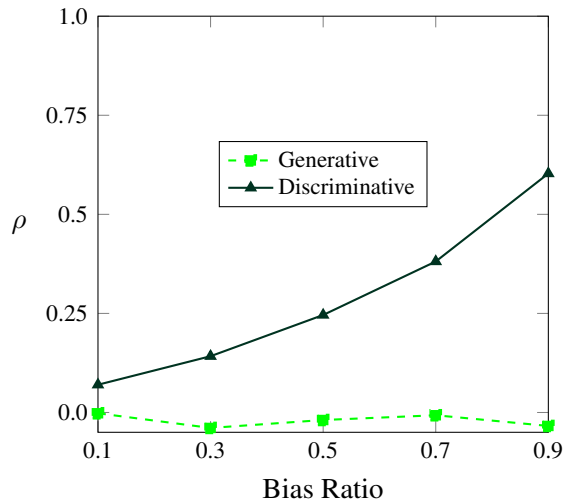

\subsection{Ablation study of fine-tuning pipeline} \label{app:ablations}

Our fine-tuning pipeline allows different ways to combine the steps, such as choosing a prior or whether to use another step of fine-tuning. 
Table~\ref{table:ablations} presents an ablation study of the different possible combinations, using BART on SNLI with hypothesis-only bias.
(The table shows means and standard deviations of 3 runs with different random seeds.)
Row 1 shows the results of the generative model, without any fine-tuning; the same model from Section~\ref{section:disc2gen}. 
Fine-tuning with a hypothesis-only prior leads to a smaller gap than fine-tuning with a uniform prior (compare rows 3 and 5). 
We can explain these apparently surprising results by the hypothesis-only prior capturing some of the bias, such that removing it during inference allows the predictions to be less biased. 
Fine-tuning with a uniform prior does not allow such a decomposition, resulting in a large gap (row 3). 
In contrast, using a hypothesis-only prior at inference leads to biased predictions (large generalization gaps; rows 2, 4 and 6). These models perform well on the test set (relative to using uniform prior at inference; rows 3, 5, 7), but relatively poorly on the o.o.d set. 
In fact, maintaining the same kind of prior throughout the pipeline (rows 3 and 6) leads to results similar to the discriminative baseline (row 11).

\begin{table*}[t]
    \centering
    {
    \begin{tabular}{c c c c r r r}
    \toprule
     & & \multicolumn{2}{c}{\textbf{Prior}}  \\ 
    \cmidrule(lr){3-4}
     & \textbf{Training} & \textbf{Fine-tuning} & \textbf{Inference} & \multicolumn{1}{c}{\textbf{Dev}} & \multicolumn{1}{c}{\textbf{Hard Dev}} & \multicolumn{1}{c}{\textbf{$\Delta$}}\\
    \midrule
    1 & \multirow{6}{*}{\rotatebox[origin=c]{0}{Gen.}} & -- & Uniform & $71.14\pm 0.4$ & $72.68\pm 0.2$ & $\boldsymbol{-1.54}\pm 0.3\phantom{0}$ \\
    2 & {} & -- & Hypothesis-only & $84.99\pm 0.3$ & $64.29\pm 0.7$ & $20.69\pm 0.9\phantom{0}$ \\
    3 & {} & Uniform & Uniform & $\boldsymbol{90.32}\pm 0.1$ & $77.17\pm 0.8$ & $13.15\pm 0.7\phantom{0}$ \\
    4 & {} & Uniform & Hypothesis-only & $89.31\pm 0.5$ & $70.33\pm 1.9$ & $18.98\pm 1.4\phantom{0}$\\
    5 & {} & Hypothesis-only & Uniform & $87.12\pm 0.5$ & $\boldsymbol{80.55}\pm 0.1$ & $6.57\pm 0.5$\textbf{*} \\
    6 & {} & Hypothesis-only & Hypothesis-only & $90.06\pm 0.0$ & $75.53\pm 0.8$ & $14.53\pm 0.8\phantom{0}$ \\
    \midrule
    7 & \multirow{4}{*}{--} & Uniform & Uniform & $90.05\pm 0.1$ & $76.54\pm 0.4$ & $13.51\pm 0.3\phantom{0}$ \\
    8 & {} & Uniform & Hypothesis-only & $89.66\pm 0.4$ & $71.71\pm 1.5$ & $17.95\pm 1.2\phantom{0}$ \\
    9 & {} & Hypothesis-only & Uniform & $87.11\pm 0.9$ & $80.53\pm 1.2$ & $6.58\pm 0.6\phantom{0}$ \\
    10 & {} & Hypothesis-only & Hypothesis-only & $90.04\pm 0.2$ & $76.13\pm 0.6$ & $13.91\pm 0.4\phantom{0}$ \\
    \midrule
    11 & \multicolumn{3}{c}{Discriminative baseline} & $91.49\pm 0.0$ & $79.59\pm 0.5$ & $11.90\pm 0.5\phantom{0}$ \\
    \bottomrule
    \end{tabular}
    }
    \caption{Ablations on SNLI validation set (Dev) with BART-base. Hard Dev was created similarly to SNLI hard \citep{gururangan2018annotation}. Fine-tuning is done with a discriminative objective, while inference is always using the generative objective. Uniform/Hypothesis-only refers to the kind of prior that was used during this phase. %
    ``*'' marks the model with the smallest o.o.d generalization gap ($\Delta$) that is better than the discriminative baseline.}
    \label{table:ablations}
    \end{table*}
 
 The fine-tuning step allows a balancing of bias and performance. Fine-tuning with a hypothesis-only prior and using a uniform prior at test time results in good o.o.d performance and relatively small generalization gaps (row 5). This setting achieve the smallest generalization gap that still beats the discriminative baseline (row 11). 
 
 Another consideration is the additional training time incurred by two phases of training. If we skip the generative training phase and directly train with the discriminative objective, we lose a bit in terms of test performance but maintain a good o.o.d performance, resulting in a medium-size generalization gap (row 9).
 
 The model on row 9 shows comparable performance to the one in row 5, with a slight performance drop and a larger standard derivation. In practice, that model demonstrated slight instability and performed worse on the test and hard test sets than the model on row 6, showing that the initial generative training phase may allow the model to generalize better.

\begin{table*}[t]
    \centering
    \begin{tabular}{l c c c c}
    \toprule
        \textbf{Model} & \textbf{Learning rate} & \textbf{No. of epochs} & \textbf{Word dropout} & \textbf{Weight decay} \\
    \midrule
        \begin{tabular}{@{}l@{}}Discriminative /\\ Hypothesis-only\end{tabular} & $10^{-5}$ & $20$ & -- & --\\
        Generative & $10^{-5}$ & $20$ & -- & --\\
        Fine-tuning & $5\cdot 10^{-6}$& $5$ & $0.1$ & $0.1$\\
    \bottomrule
    \end{tabular}
    \caption{Hyperparameters for models. All of the models used early stooping of $3$ epochs without improvement.}
    \label{table:hyperparams}
\end{table*}

\subsection{Hyperparameters and Training Details}
Table~\ref{table:hyperparams} shows the hyperparameters for the models used throughout the paper. 
We experimented with word dropout values of: $0.01,0.1,0.3,0.5$, weight decay values of: $0.001,0.01,0.1,1$, learning rate values in the range: $[10^-6,10^-4]$, and maximum number of $5,10,20$ and $100$ epochs. The values that achieved the best accuracy on the validation set appear in the table. %
All other hyperparameters are the default ones in~\citet{wolf2019huggingface}. 
Where mean and standard deviation is specified, we calculate those values over 3 runs, each with a different seed. Otherwise, those are the results of only one run.

Each experiment was performed on one or two NVIDIA RTX 2080 Ti GPUs.
Training takes about 6--7 hours for discriminative models, 7--8 hours for generative models, and 15--20 hours for the discriminative fine-tuning step. 
Discriminative BERT/BART models have 109M/140M %
parameters, while generative BERT/BART models have 247M/139M %
parameters.

\end{document}